# Multi-UAV Swarm Obstacle Avoidance Based on Potential Field Optimization

Yendo Hu [1*], Yiliang Wu [1], Weican Chen [1]

[1]College of Marine Information Engineering, Jimei University, Xiamen, China

*Corresponding author:201861000005@jmu.edu

**Abstract.** In multi-UAV scenarios, the traditional Artificial Potential Field (APF) method often leads to redundant flight paths and frequent abrupt heading changes due to unreasonable obstacle avoidance path planning, and is highly prone to inter-UAV collisions during the obstacle avoidance process. To address these issues, this study proposes a novel hybrid algorithm that combines the improved Multi-Robot Formation Obstacle Avoidance (MRF-IAPF) algorithm [1] with an enhanced APF optimized for single-UAV path planning. Its core ideas are as follows: first, integrating three types of interaction forces from MRF-IAPF—obstacle repulsion force, inter-UAV interaction force, and target attraction force; second, incorporating a refined single-UAV path optimization mechanism, including collision risk assessment and an auxiliary sub-goal strategy. When a UAV faces a high collision threat, temporary waypoints are generated to guide obstacle avoidance, ensuring eventual precise arrival at the actual target. Simulation results demonstrate that compared with traditional APF-based formation algorithms, the proposed algorithm achieves significant improvements in path length optimization and heading stability, can effectively avoid obstacles and quickly restore the formation configuration, thus verifying its applicability and effectiveness in static environments with unknown obstacles.

## 1. Introduction

In recent years, the application scope of unmanned aerial vehicles (UAVs) has been expanding rapidly[2], and they have been widely used in various fields such as military reconnaissance, emergency rescue, and environmental conservation[3]. As a core key technology for UAV flight control, path planning can ensure the stable flight of UAVs and safely and efficiently avoid various obstacles[4]. Current mainstream path planning methods include the A* algorithm[5], genetic algorithm[6], particle swarm optimization algorithm[7], ant colony optimization algorithm[8], rapidly exploring random tree (RRT) algorithm[9], velocity obstacle method[10,11], reinforcement learning methods[12], and artificial potential field (APF) method[13-21], etc.

Among numerous path planning methods, the artificial potential field (APF) method has gained extensive attention and application in the UAV field due to its significant advantages such as low computational cost, fast real-time response, simple control logic, and strong robustness[22]. However, the traditional APF method still has inherent limitations, which may lead to unreasonable obstacle avoidance path planning. In response to this situation, scholars at home and abroad have carried out a large number of improved studies. Feng et al.[14] incorporated the velocity constraints of the aircraft into the repulsion force calculation model and quantitatively classified the threat levels of obstacles to improve the scientificity of obstacle avoidance decisions. Yang et al.[15] optimized and adjusted the gravitational model to enhance the traction effect of the target point on the UAV, and introduced auxiliary target points and an evaluation mechanism to effectively reduce the probability of the UAV falling into local minima. Nevertheless, this method does not dynamically set auxiliary sub-targets based on the relative position information between the UAV and obstacles, which affects the flexibility of obstacle avoidance.

In the direction of path length optimization, Jiang et al.[20] proposed an adaptive step size adjustment strategy aimed at shortening the redundant obstacle avoidance paths generated by the traditional APF method; Li et al.[21] introduced the ratio of the total path length to the distance from the current position to the target point into the path calculation process, realizing the dynamic adaptive adjustment of the step size. Both methods focus on improving path efficiency and provide effective ideas for optimizing the path planning performance of the APF method.

The main contributions of this paper are as follows:

(1) By considering the Euclidean distance between UAVs, the interaction velocity among UAVs and the k-nearest neighbor search mechanism are introduced to achieve inter-UAV collision avoidance, as well as the rapid formation and stable maintenance of the formation.

(2) An improved hybrid APF algorithm is proposed. The repulsive potential field function is optimized to ensure its first-order gradient continuity, and a collision risk assessment mechanism and an auxiliary sub-target strategy are integrated. This guides the UAV swarm to complete obstacle avoidance along a more reasonable path, effectively shortening the flight path length and reducing the number of abrupt heading angle changes.

## 2. Theoretical Model

In this section, a multi-UAV system model based on the Artificial Potential Field (APF) method is constructed, and the theoretical basis and mathematical modeling process related to its formation maintenance and obstacle avoidance control are systematically elaborated.

### 2.1 Formation

Based on graph theory, the multi-UAV system can be mapped to an undirected graph $G(q,w)$, Herein, $q = \{q_1, q_2 \ldots q_n\}$ denotes a non-empty finite set, where each element corresponds to a single UAV. $w \subseteq q \times q$ represents the set of edges between UAVs, and the adjacency matrix $A = [a_{ij}] \in R^{n \times n}$ is used to describe the adjacency relationship among different UAVs. If $a_{ij} \neq 0$, it indicates that $(i,j) \in w$, meaning UAV $R_i$ and $R_j$ are capable of mutual perception [23, 24]. For each UAV, its neighbor set is defined as:

$$N_i^R = \{j \in q: a_{ij} \neq 0\} = \{j \in q: (i,j) \in w\} \tag{1}$$

The communication radius of each UAV is defined as $r$. As shown in Fig 1, the neighbors of a UAV are defined as other UAVs within a circle centered at the UAV's current position with radius $r$. UAVs connected by solid lines satisfy the distance constraint and thus form a neighbor relationship, while those connected by dashed lines do not meet the distance constraint and therefore do not constitute a neighbor relationship. The specific mathematical definition is as follows:

$$N_i = \{j \in q: ||P_j - P_i|| < r\} \tag{2}$$

Where $P$ denotes the current position of the UAV, and $||P_j - P_i||$ represents the Euclidean distance between UAV $j$ and UAV $i$.

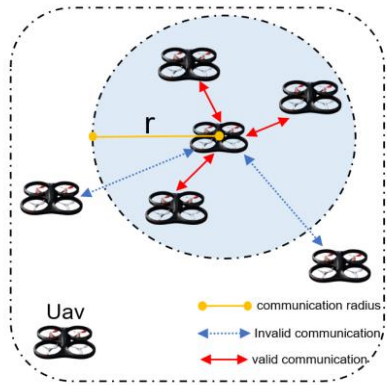

Fig. 1 Multi-UAV Communication Model

Fig. 2 presents the cluster model constructed based on the communication model in Fig. 1. During the cluster cooperative operation of multi-UAVs, each UAV in the formation needs to maintain a preset equal distance, and thus the following constraint condition must be satisfied:

$$d - \varphi < \left\| P_j - P_i \right\| < d + \varphi, \forall (i,j) \in w(P) \tag{3}$$

Where $d$ denotes the preset ideal distance between UAVs, and $\varphi$ represents the allowable distance deviation threshold. The construction of this cluster model is based on the communication model in Fig. 1 and the above-mentioned distance constraint condition.

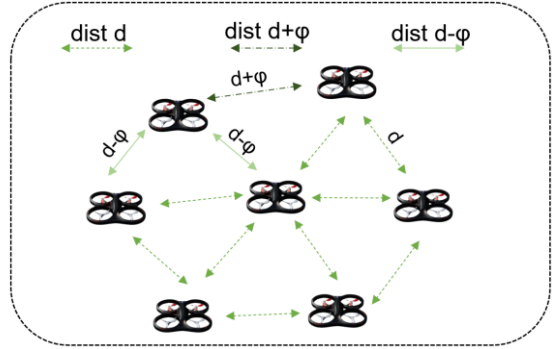

Fig. 2 Multi-UAV Cluster Model

In this study, the Artificial Potential Field (APF) function is adopted to achieve the formation and stable maintenance of the target distance between multi-UAVs. The interaction force between neighboring UAVs defined by the APF function is as follows:

$$\left\| P_j - P_i \right\| = dist, \forall (i,j) \in w(P) \tag{4}$$

$$F_{int} = \begin{cases} -\beta \left( \dfrac{1}{dist} - \dfrac{1}{d} \right) \left( \dfrac{1}{dist} \right)^2 \overrightarrow{dist}, & dist < d \\ 0, & dist = d \\ \alpha \left( \dfrac{2}{1 + e^{-dist}} - 1 \right) (dist - d) \overrightarrow{dist}, & dist > d \end{cases} \tag{5}$$

When the current distance between UAVs is less than the ideal distance $d$, the interaction force acts as a repulsive force; when the distance equals $d$, the interaction force is 0; when the distance is greater than $d$, the interaction force acts as an attractive force. Herein, $\beta$ and $\alpha$ are positive real coefficients, which are used to adjust the intensities of the repulsive force and attractive force.

**2.2 Obstacle Avoidance**

The core idea of the Artificial Potential Field (APF) is to model the UAV's operating environment as a virtual artificial potential field: the target point generates an attractive potential field, while various obstacles produce repulsive potential fields. By superimposing the attractive and repulsive potential fields to form a total potential field, the UAV can autonomously plan its flight path under the influence of this total field. After incorporating the neighbor interaction force for formation (mentioned above), the combined force acting on a single UAV is illustrated in Fig. 3.

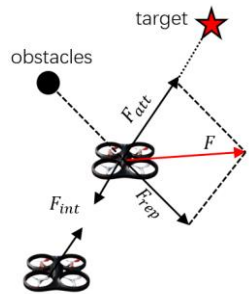

Fig. 3 UAV Total Force

In the traditional Artificial Potential Field (APF) method, the potential function of the attractive potential field is defined as follows:

$$U_{att}(p) = \frac{1}{2}k_{att}d^2(p,p_t) \tag{6}$$

The attractive force is the negative gradient of the attractive potential field, so the expression for the attractive force $F_{att}(p)$ is:

$$F_{att}(p) = -\nabla U_{att}(p) = -k_{att}d(p,p_t)\vec{d_t} \tag{7}$$

where $k_{att}$ is a positive real attractive coefficient, $p$ denotes the current position of the UAV, $p_t$ denotes the position of the target point, $d(p,p_t)$ is the Euclidean distance between the UAV and the target point, and $\vec{d_t}$ is the unit vector pointing from the UAV to the target point.

The repulsive potential field function of the traditional APF is defined in a piecewise form:

$$U_{rep}(p) = \begin{cases} \frac{1}{2}k_{rep}\left(\frac{1}{d(p,p_o)} - \frac{1}{d_o}\right)^2 & d(p,p_o) \leq d_o \\ 0 & d(p,p_o) > d_o \end{cases} \tag{8}$$

The repulsive force is the negative gradient of the repulsive potential field, and similarly, the expression for the repulsive force $F_{rep}(p)$ is:

$$F_{rep}(p) = \begin{cases} k_{rep}\left(\frac{1}{d(p,p_o)} - \frac{1}{d_o}\right)\frac{1}{d^2(p,p_o)}\vec{d_o} & d(p,p_o) \leq d_o \\ 0 & d(p,p_o) > d_o \end{cases} \tag{9}$$

where $k_{rep}$ is a positive real repulsive coefficient, $p_o$ denotes the position of the obstacle, $d(p,p_o)$ is the Euclidean distance between the UAV and the obstacle, and $d_o$ is a positive real obstacle avoidance radius. The repulsive force only takes effect when the distance between the UAV and the obstacle is less than the obstacle avoidance radius; $\vec{d_o}$ is the unit vector pointing from the obstacle to the UAV.

If the UAV is subjected to n repulsive forces generated by n obstacles, the total resultant force $F(p)$ acting on the UAV can be expressed as:

$$F(p) = F_{att}(p) + \sum_{i=1}^{n} F_{rep_i}(p) \tag{10}$$

When applying the traditional Artificial Potential Field (APF) method in multi-UAV formation flight scenarios, it is necessary to simultaneously consider the influence of interaction forces between UAVs and the relative relationship between the current flight trajectory and obstacles (i.e., whether obstacle avoidance operations need to be performed). Restricted by the aforementioned factors, the traditional APF method has the following key issues:

(1) Interaction forces between UAVs are prone to causing flight trajectory distortion;

(2) Unreasonable maneuvering behaviors may occur for obstacle avoidance.

Both types of problems will have a negative impact on the path planning effect of the multi-UAV formation.

To improve the aforementioned problems, this paper proposes targeted improvements to the traditional Artificial Potential Field (APF) method: to tackle the unreasonable maneuvering behaviors that may occur during obstacle avoidance, a collision risk assessment mechanism based on the safe distance angle is introduced; to address the collision risks faced by UAVs during flight or the interference of inter-UAV interaction forces, an auxiliary sub-target strategy is designed. By guiding UAVs to avoid obstacles accurately and weakening the negative impacts caused by inter-UAV interaction forces.

2.2.1 Obstacle Detection and Localization

Obstacle detection and localization is a prerequisite for UAV obstacle avoidance. Considering lightweight design, traditional image processing methods are adopted in this study. The process starts with acquiring depth (perspective/planar) and segmentation images: depth images provide pixel-wise distance information, while segmentation images are used to label obstacle regions. Subsequently, the 3D coordinates $(x, y, z)$ of each pixel are solved using depth images. Finally, connected components are extracted from segmentation images, their spatial positions are obtained via the pixel 3D coordinate solving module, and valid obstacles are filtered by the distance threshold $dis_{threshold}$ with their 3D bounding boxes stored. This process is visualized in the flow diagram (Fig. 4).

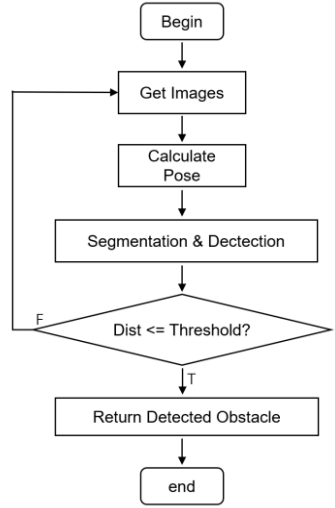

Fig. 4 Obstacle Detection and Localization Flow Diagram

2.2.2 Mechanism for Calculating Collision Risk Based on Safe Distance Angle

On the basis of the traditional APF, a mechanism for calculating collision risk based on the angle $\varphi$ of the UAV's current flight path is added as shown in Fig. 5. Here, $\theta_L$ and $\theta_R$ are two boundaries. The cross product of $-\vec{d_o}$ and $\vec{d_t}$ is calculated to determine which boundary to use, and $\theta$ is a redundant term for linear division. If the cross product of $-\vec{d_o}$ and $\vec{d_t}$ is greater than 0, the boundary is $\theta_L$; otherwise, the boundary is $\theta_R$. The following formula (11) shows the case where the boundary is $\theta_L$:

$$R_{col} = \begin{cases} 1.0 - \dfrac{\varphi - \theta}{\theta_L - \theta} & \theta < \varphi < \theta_L \\ 1.0 & \varphi < \theta \end{cases} \quad (11)$$

Fig. 5 also shows the positions of target points with risk values of 0, 1, or a value between 0 and 1, where different risk values are marked with distinct colors. To avoid cluttering the figure, the specific calculation process of the risk value for one UAV is elaborated in detail, while the relevant illustration for the other UAV is simplified.

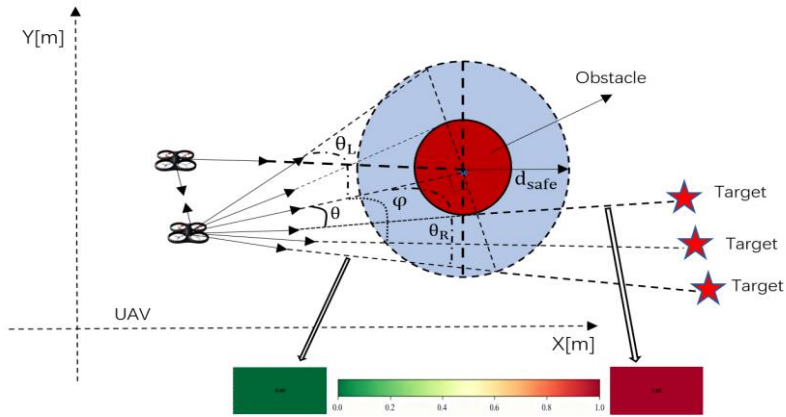

Fig. 5 Obstacle Collision Risk Mechanism

### 2.2.3 Auxiliary Sub-Targets Generate

In multi-UAV formation flight scenarios, when there is a collision risk on the UAV's forward flight trajectory and obstacle avoidance needs to be performed, the traditional Artificial Potential Field (APF) method will cause an abrupt change in the obstacle avoidance turning angle due to the sudden increase in repulsive force, thereby affecting flight stability and smoothness. Meanwhile, restricted by their actual turning performance, UAVs cannot achieve large-scale abrupt angle changes [25,26]. To address this, this paper introduces an auxiliary sub-target strategy, which not only effectively reduces the magnitude of angle changes in the flight path and ensures flight smoothness but also helps UAVs avoid obstacles accurately or weaken the interference of inter-UAV interaction forces, ultimately meeting the requirements of actual flight performance.

$$x_{aux_i} = x_{obs} \pm \sqrt{\frac{d_{safe}^2}{1+\left(\frac{-1}{k_1}\right)^2}} \ (i = 1,2), \tag{12}$$

$$y_{aux_i} = y_{obs} \pm \sqrt{\frac{d_{safe}^2}{1+\left(\frac{-1}{k_1}\right)^2}} \ (i = 1,2), \tag{13}$$

As shown in Fig. 6, with the UAV's current position $P_{UAV\_cur}(x_n, y_n)$ as the origin, the detection range is a 90° fan-shaped area ahead (radius $d_{pre}$). When obstacle $P_{obs}(x_{obs}, y_{obs})$ is detected: if the collision risk is below the threshold, the obstacle's repulsive force is 0; if above, auxiliary sub-targets $P_{aux\_i}(i = 1,2)$ are generated to guide flight. These targets are obtained by the intersection of perpendicular line $L_2$, obstacle line $L_1$ (slope $k_1$) and safe distance $d_{safe}$, with coordinates given in Equations (12) and (13).

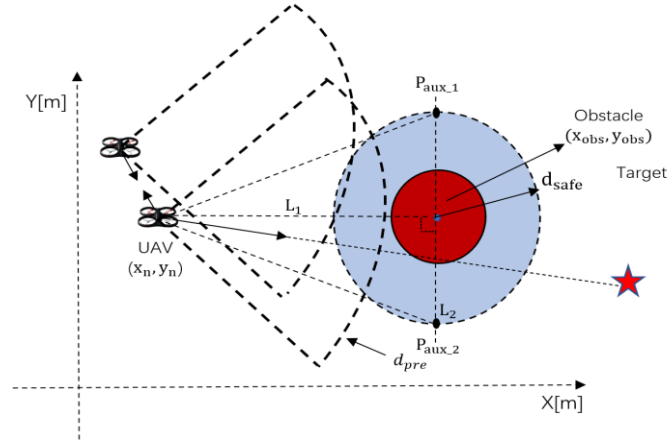

Fig. 6 Auxiliary Sub-Targets Generate

### 2.2.4 Artificial Potential Field Improvement

The attractive force in the traditional Artificial Potential Field (APF) method depends solely on the distance between the UAV and the target point: as the UAV approaches the target, the attractive force gradually diminishes, which may hinder precise arrival at the target point. To address this issue, this paper improves the calculation of the target attractive force by introducing the ratio of the distance between the UAV's starting point and the target point $||P_s - P_t||$ to the distance between its current position and the target point $||P_{UAV\_cur} - P_t||$. The enhanced attractive force $F_{en\_att}(target)$ after improvement can provide a stronger traction effect when the UAV is close to the target point, ensuring precise arrival, and its expression is given in Equation (14).

$$F_{en_{att}}(target) = k_{att} e^\rho \tag{14}$$

The parameter $\rho$ satisfies the equation (15), $\gamma$ is the adjustment coefficient:

$$\rho = \frac{\|P_s - P_t\|^\gamma}{\frac{\|P_s - P_t\|}{2} + \|P_{UAV_{cur}} - P_t\|} \tag{15}$$

When a auxiliary sub-target exists, the attraction is generated by the auxiliary sub-target. The UAV moves to the next position due to the attraction of the auxiliary sub-target towards the goal point. The equation for the attraction force $F_{aux\_att}(target\_aux)$ of the auxiliary sub-target is given by equation (16). The equation is derived based on the distance $\|P_{UAV\_cur} - P_{aux}\|$ between the UAV and the auxiliary sub-target.

$$F_{aux_{att}}(target_{aux}) = k_{att} e^{\left(\frac{\delta}{\|P_{UAV_{cur}} - P_{aux}\|}\right)^2} \tag{16}$$

$\delta$ is the distance parameter in the equation. Equation (17) provides the expression for the attractive force $F_{att}$ exerted on the UAV, following the conditions in equations (14) and (16).

$$F_{att} = \begin{cases} F_{en_{att}}(target) &, \nexists(target_{aux}) \\ F_{aux_{att}}(target_{aux}), & \exists(target_{aux}) \end{cases} \tag{17}$$

## 3. Experiment

Since the effectiveness of some of the aforementioned mechanisms has been verified in Reference [27] for scenarios with known static obstacles and a single UAV, this paper mainly aims to verify their performance in optimizing the obstacle avoidance paths of multi-UAV swarms in scenarios with unknown static obstacles. To validate the above performance, a static obstacle scenario is constructed in AirSim, and a comparison is made between the traditional artificial potential field method and the improved artificial potential field method proposed in this paper.

Among them, the indicators for performance comparison are defined as follows:

$$P_{length} = \sum_{i=1}^{m-1} \sqrt{(x_{n+1} - x_n)^2 + (y_{n+1} - y_n)^2}$$

The total path length of the UAV swarm flight is defined as shown in the above formula, where m represents the total number of flight steps of the UAVs and n represents the number of UAVs.

$$A_{change} = \frac{180° \times arctan\left(\frac{y_{n+1} - y_n}{x_{n+1} - x_n}\right)}{\pi}$$

$$A_{change\_count} = \sum_{i=1}^{n-1} I(A_{change} > 5°)$$

The formulas for the angle change and the number of angle changes in the UAV's flight path are as shown above. Here, $I$ is an indicator function: when the expression inside the parentheses is true, the count increases by 1; otherwise, the count increases by 0.

The following table presents the specific parameter values used in the aforementioned formulas.

Table1 Parameters used in the experiment

| Parameter | Value |
|---|---|
| $r$ | 6 |
| $d$ | 4 |
| $\varphi$ | 0.7 |
| $\alpha$ | 0.1 |
| $\beta$ | 10 |
| $k_{att}$ | 1 |

| $k_{rep}$ | 500 |
|---|---|
| $d_{pre}$ | 6 |
| $\gamma$ | 1.2 |
| $\delta$ | 16 |
| $dis_{threshold}$ | 6 |

A simulation scenario is constructed as follows: the total number of UAVs is 5, and its top view is shown in Fig. 7.

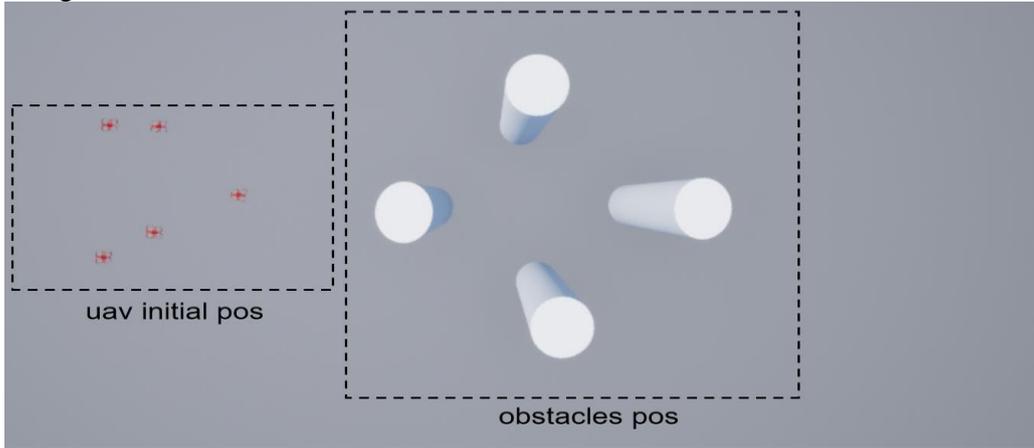

Fig. 7 Top View of The Simulation Scenario

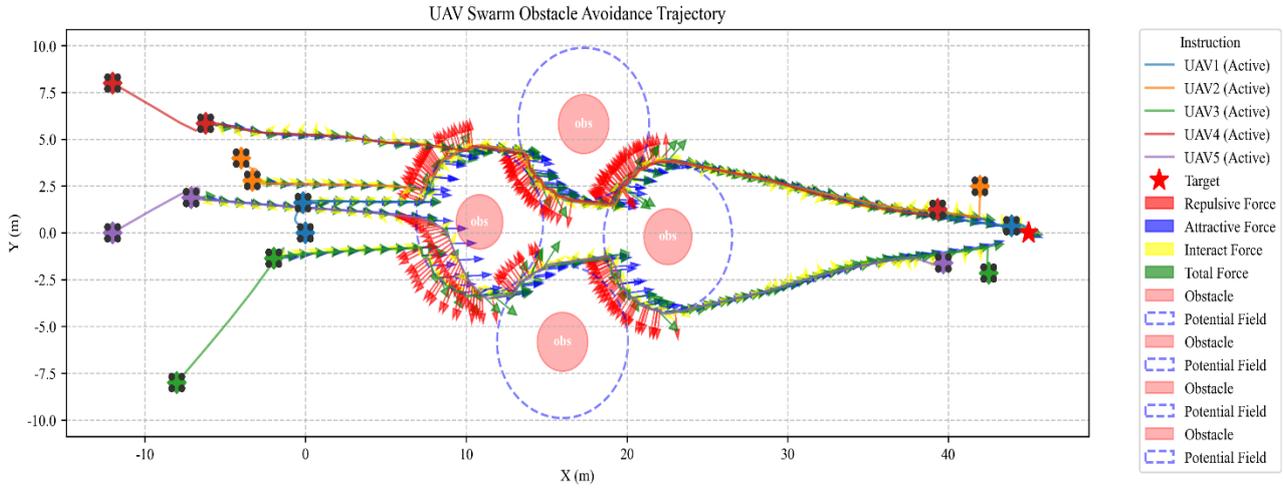

Fig. 8.a UAV Swarm Obstacle Avoidance Trajectory(T-APF)

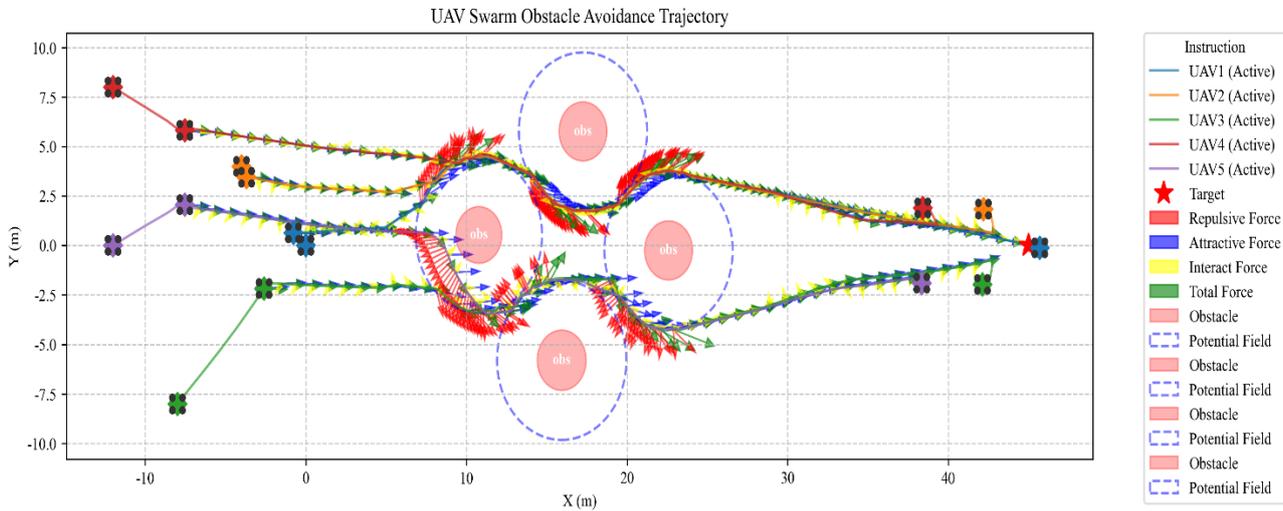

Fig. 8.b UAV Swarm Obstacle Avoidance Trajectory(I-APF)

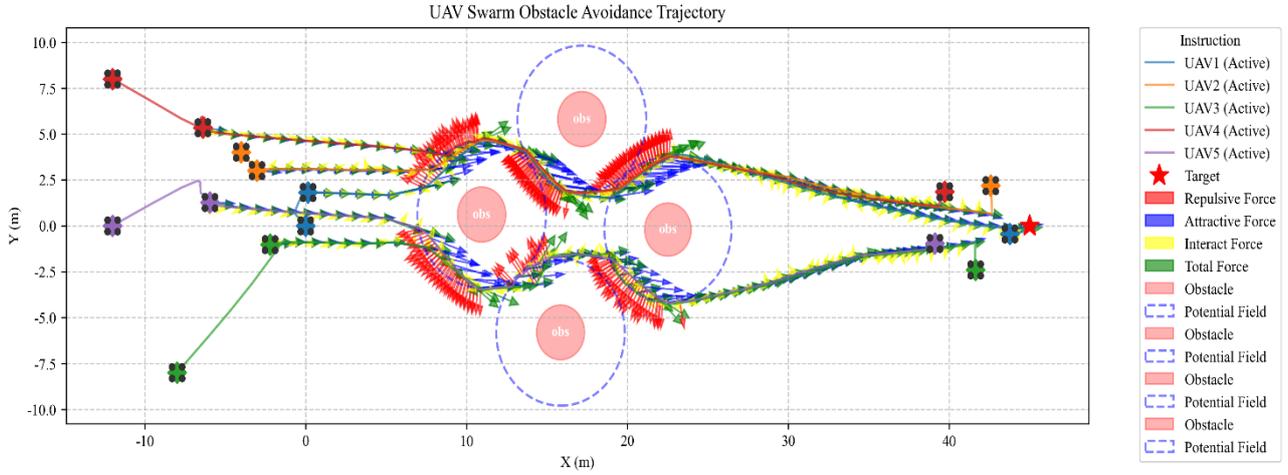

Fig. 8.c UAV Swarm Obstacle Avoidance Trajectory(O-APF)

Fig. 8.a, 8.b, and 8.c present the comparative results of path planning for five UAVs in the scenario illustrated in Fig. 7, using the traditional Artificial Potential Field (T-APF) method, the improved method from Reference [28] (I-APF), and the hybrid method proposed in this paper (O-APF), respectively. It can be observed that compared with T-APF, both I-APF and O-APF significantly improve the smoothness of the obstacle avoidance path, indicating that the auxiliary sub-goal generation mechanism integrated in the two algorithms exerts a positive effect on mitigating unreasonable obstacle avoidance behaviors during multi-UAV formation obstacle avoidance. However, I-APF fails to alleviate the unreasonable obstacle avoidance caused by inter-UAV interaction forces, while O-APF effectively addresses this issue. This result verifies that with the synergy of the auxiliary sub-goal generation mechanism, optimizing the attractive potential field yields a more remarkable effect on enhancing the obstacle avoidance performance of multi-UAV formations than merely improving the repulsive potential field.

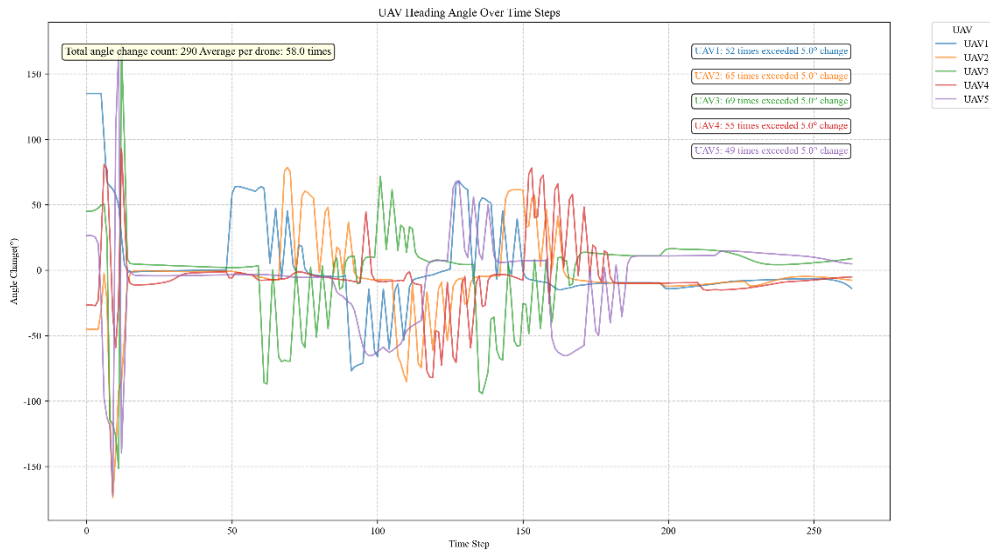

Fig. 9.a UAV Heading Angle(T-APF)

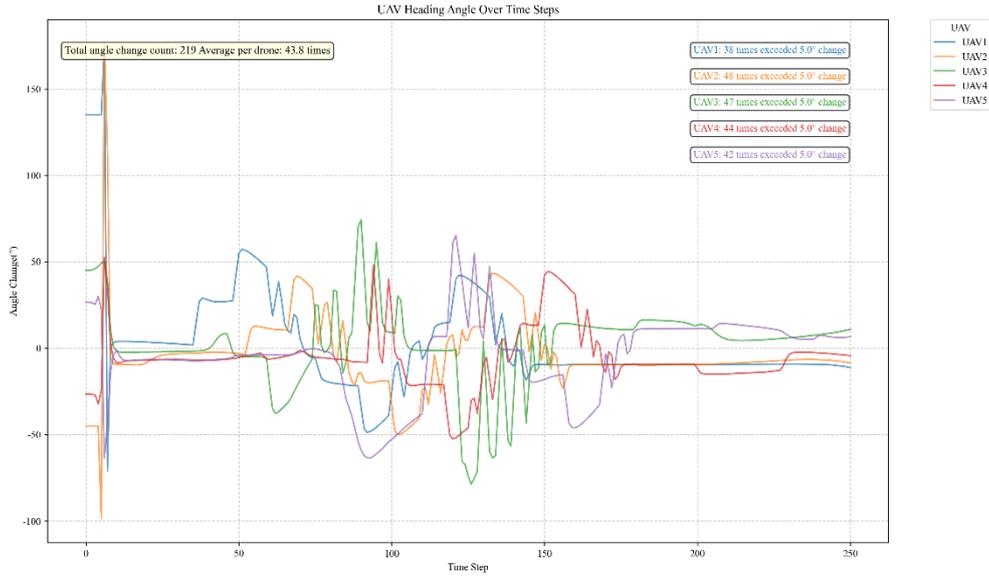

Fig. 9.b UAV Heading Angle(I-APF)

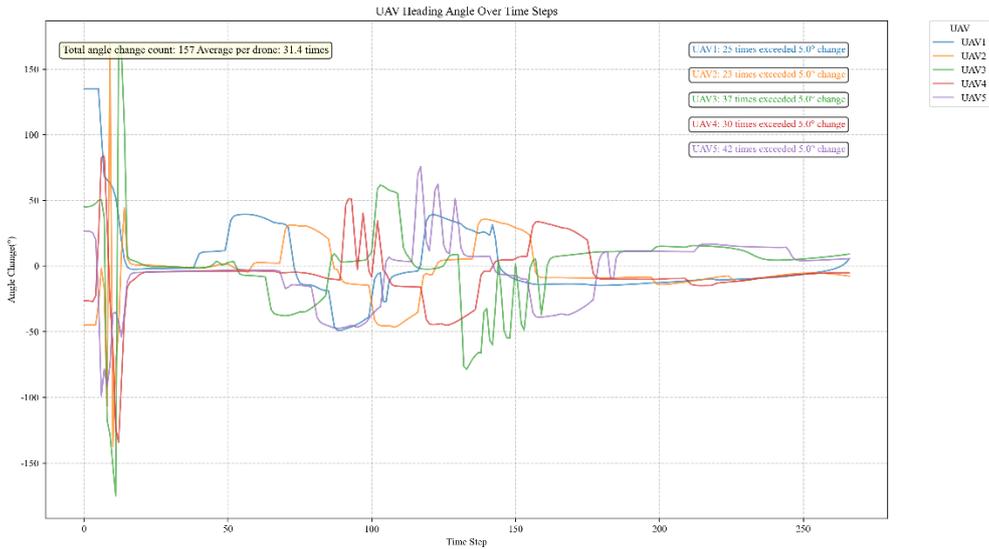

Fig. 9.c UAV Heading Angle(O-APF)

Figures 9.a, 9.b, and 9.c compare the number of heading angle changes of UAVs corresponding to the paths in Figures 8.a, 8.b, and 8.c, respectively. As can be clearly seen from the data in the figures: compared with Figure 9.a (T-APF), the number of heading angle changes of UAVs in Figure 9.b (I-APF) decreases by 32%, while that in Figure 9.c (O-APF) drops by as much as 84%. This quantitative result further confirms that the O-APF proposed in this paper is significantly superior to I-APF in optimizing path smoothness.

Table2 Indicator Comparison Table(UAV(1/3/5))

| Algorithm | Path Length [m] | Rate | Angle change[N] | Rate |
|---|---|---|---|---|
| T-APF | 49.26/307.27/805.56 |  | 98/214/290 |  |
| I-APF | 47.58/302.61/786.36 | 3.5%/1.5%/2.4% | 60/162/219 | 63%/32%/32% |
| O-APF | 46.18/300.31/774.53 | 6.6%/2.3%/4.0% | 18/82/157 | 440%/160%/84% |

As shown in the above table, it compares the performance of the three algorithms in terms of path length and number of heading angle changes under different numbers of UAVs. It can be observed that with the increase in the number of UAVs, compared with the traditional T-APF, the improvement margins of both I-APF and O-APF for the above two indicators show a declining trend, and the attenuation of path length is not linear. However, overall, the two improved algorithms still maintain a performance advantage over T-APF, and the O-APF proposed in this paper is superior to I-APF in all indicators.

## 4. Conclusion

This paper first systematically combs through the core theoretical models in the field of multi-UAV formation and obstacle avoidance. By conducting an in-depth analysis of the inherent flaws of the traditional Artificial Potential Field (T-APF) method, it clarifies the problems such as unreasonable paths and inter-UAV interaction interference existing in multi-UAV cooperative obstacle avoidance scenarios. The existing algorithm is improved by introducing a collision risk assessment mechanism and auxiliary sub-goal generation, and scenarios are constructed in AirSim to compare the traditional APF (T-APF), the method in Reference [28] (I-APF), and the hybrid APF (O-APF) proposed in this paper, verifying the effectiveness of the improved APF.

Simulation experiments show that the O-APF effectively improves the problems of unreasonable obstacle avoidance and irrational UAV paths caused by inter-UAV interaction forces in multi-UAV obstacle avoidance, and makes the path smoother compared with the traditional APF.

In the future, we will further study potential fields and optimize the attractive potential field and repulsive potential field to make the distribution of potential fields more suitable for UAV path planning and obstacle avoidance. In addition, we will further optimize the auxiliary target point obstacle avoidance strategy to improve the obstacle avoidance efficiency of robots and shorten the obstacle avoidance path.